\documentclass{article}
\usepackage{spconf,amsmath,graphicx,hyperref}

\usepackage{microtype}

\usepackage{booktabs}  
\usepackage{arydshln}

\usepackage{multirow}
\usepackage{float}
\usepackage{mismath}
\usepackage[many]{tcolorbox}
\usepackage{amssymb}
\usepackage{mathtools}
\usepackage{amsthm}
\usepackage{pifont}
\usepackage{enumitem}
\usepackage{CJKutf8}



\newcommand{\keypoint}[1]{\noindent \textbf{#1}:}

\usepackage{subcaption}

\usepackage{algorithm}
\usepackage[noend]{algpseudocode}

\usepackage[english]{babel} 

\usepackage{tikz}
\usetikzlibrary{decorations.pathreplacing,calc}

\tikzset{brace/.style={decorate, decoration={brace}},
 brace mirrored/.style={decorate, decoration={brace,mirror}},
}

\newcounter{brace}
\newcounter{arrow}
\title{Clustering-driven Memory Compression for\\ On-device Large Language Models}
\name{Ondrej Bohdal\textsuperscript{1}, Pramit Saha\textsuperscript{1,2,*}\thanks{\textsuperscript{*}Work done during an internship at Samsung R\&D Institute UK.}, Umberto Michieli\textsuperscript{1}, Mete Ozay\textsuperscript{1}, Taha Ceritli\textsuperscript{1}}
\address{\textsuperscript{1}Samsung R\&D Institute UK \ \textsuperscript{2}University of Oxford}

\begin{document}
\maketitle
\begin{abstract}
Large language models (LLMs) often rely on user-specific memories distilled from past interactions to enable personalized generation. A common practice is to concatenate these memories with the input prompt, but this approach quickly exhausts the limited context available in on-device LLMs. Compressing memories by averaging can mitigate context growth, yet it frequently harms performance due to semantic conflicts across heterogeneous memories. In this work, we introduce a clustering-based memory compression strategy that balances context efficiency and personalization quality. Our method groups memories by similarity and merges them within clusters prior to concatenation, thereby preserving coherence while reducing redundancy. Experiments demonstrate that our approach substantially lowers the number of memory tokens while outperforming baseline strategies such as naive averaging or direct concatenation. Furthermore, for a fixed context budget, clustering-driven merging yields more compact memory representations and consistently enhances generation quality.
\end{abstract}
\begin{keywords}
Personalization, memory, large language models, on-device, efficiency.
\end{keywords}

\section{Introduction}

Large language models (LLMs) have demonstrated remarkable performance across a wide range of natural language processing tasks \cite{zhao2023survey, minaee2024large}, including question answering \cite{sticha2024qa}, text summarization \cite{liu2023learning}, rewriting \cite{shu2024rewritelm}, translation \cite{zhu2023multilingual}, and grammar correction \cite{severyn2021grammar}. Many of these tasks require inferring missing context from user background or prior interactions, making personalization a critical component of effective generation \cite{zhang2025personalization}. To this end, LLMs often rely on memory mechanisms that store and reuse information from past exchanges \cite{zhang2025memory}.

However, personalization via memory introduces several challenges in on-device settings, such as smartphones or edge devices \cite{dhar2021ondevice}. On-device LLMs are typically small (1–3B parameters) and therefore less capable of capturing and retaining long-term user information. In addition, their limited context windows make naive strategies—such as concatenating all relevant user histories—impractical. Effective personalization thus requires mechanisms for retrieving, filtering, and compressing user memories into compact yet informative representations.

In this paper, we propose a clustering-driven memory compression framework designed specifically for on-device personalization. Our method retrieves the most relevant textual memories, encodes them into memory tokens, and then groups related memories into clusters before merging within each cluster. This approach reduces redundancy and conflict across heterogeneous memories, thereby saving valuable context space while preserving coherence. Crucially, these savings directly translate into higher personalization quality under strict context constraints, outperforming simple concatenation or mean-based compression.

Our main contributions are as follows:
1) A clustering-based memory compression method that is both efficient and effective for on-device personalization.
2) Comprehensive evaluation across multiple datasets and LLMs, demonstrating improved performance compared to baseline approaches.
3) In-depth analyses that shed light on why clustering improves memory representations and how it balances context efficiency with personalization quality.

\begin{figure*}[t]
\vskip 0.2in
\begin{center}
\centerline{\includegraphics[width=\textwidth]{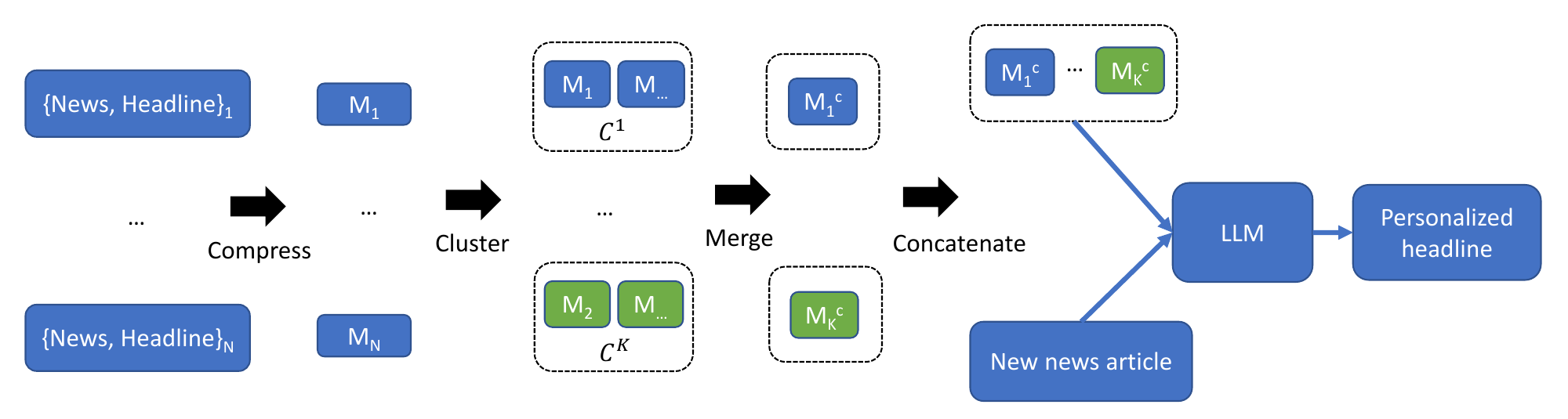}}
\caption{\textbf{Clustering-driven memory compression.} Given $N$ supporting documents (e.g., news articles with personalized headlines), we first compress each into a set of memory tokens. These memories are then grouped into $K$ clusters based on similarity. Within each cluster, the memories are merged (e.g., by averaging), and the resulting cluster-level representations are concatenated. The final compressed memories are appended to the new input example, enabling the LLM to generate personalized outputs under a limited context budget.}
\label{fig:diagram}
\end{center}
\vskip -0.2in
\end{figure*}

\section{Related Work}
\keypoint{Personalization via memory mechanism} 
To maximize the usefulness of LLMs across diverse use cases, it is essential to personalize them by maintaining user-specific information through memory \cite{zhang2025memory}. Existing approaches vary in design: some employ advanced graph-based memory systems that explicitly model interactions across memories (e.g., Mem0 \cite{mem0}, MemLoRA \cite{bini2025memlora} and MemOS \cite{li2025memos_long}), while others adopt simpler formulations where memories are compressed into synthetic prompts that are appended to the user instruction (e.g., ICAE \cite{ge2024incontext} and ComMer \cite{zeldes2025commer}). In this work, we follow the latter line, which is more suitable for on-device deployment, and represent user history through compact memory prompts.
Typically, these approaches employ a dedicated compressor—an LLM fine-tuned with parameter-efficient methods such as LoRA \cite{hu2021lora}—to encode retrieved texts into compact memory tokens. The resulting memory prompt is then concatenated with the user instruction to guide generation. A key limitation, however, is that memory tokens directly consume part of the LLM’s context window, which is particularly restrictive for on-device models (e.g., with 1,024–2,048 tokens). This motivates the need for more efficient memory compression strategies.

\keypoint{On-device LLMs}
Most state-of-the-art LLMs require substantial computational resources, such as high-end GPUs on cloud servers \cite{borzunov2024distributed}. Yet, many important use cases involve sensitive user data stored on personal devices, such as smartphones, where sending data to remote servers may be undesirable or infeasible \cite{dhar2021ondevice}. For instance, users may wish to generate personalized replies \cite{bohdal2025compositional, bohdal2025device, ceritli2025hydraopt} that leverage private information available only locally.
To address such scenarios, a growing number of compact LLMs have been developed, typically in the range of 1–3 billion parameters, that can run efficiently on-device while maintaining strong performance. Recent examples include Qwen2.5 1.5B \cite{qwen2} and Gemma3 1B \cite{team2025gemma}. However, their limited parameter count and smaller context windows make memory efficiency and compression especially critical for achieving effective personalization.

\section{Method}

For each instruction $x$ and expected response $y$ we have a collection of $N$ supporting documents $\{ \text{doc}_i \}_{i=1..N}$. The response should be personalized based on information from the $N$ documents $\{ \text{doc}_i \}_{i=1..N}$. As there may be many documents available, we first perform filtering using the BM25 method \cite{robertson2009mb25} to obtain the closest $N$ documents for the given query.

Each of the $N$ personalization documents is compressed into a memory $M_i$ via a specialized architecture \cite{ge2024incontext, zeldes2025commer}. Note that each document needs to be processed separately, as otherwise they may not fit into the available context window, a challenge for on-device LLMs. This consideration rules out approaches such as \cite{ge2024incontext} that concatenate all documents and process them within one inference. The architecture for extracting memories includes a frozen LLM with trainable embedding and LoRA \cite{hu2021lora} parameters. Each memory has $D_m$ memory tokens of dimension $D_e$, where $D_m$ is the selected number of memory tokens per document (e.g., 128) and $D_e$ is the embedding size (e.g,. 2,048).

Existing solutions \cite{zeldes2025commer} either take the mean across the $N$ memories or concatenate them to produce in total $D_m$ or $N \times D_m$ memory tokens, respectively. We use these strategies as the baselines in our evaluation. We propose an alternative approach where we cluster the memories into $K$ clusters. The resulting method decreases the total number of tokens compared to concatenation, without decreasing the performance in practice. In contrast, taking the mean decreases the performance.

We utilize K-Means clustering to cluster the memories. Within each cluster $C^k$, we take the average across the corresponding memories $\{M_j; j \in C^k\}$ to compute one clustered memory $M^c_k$ per cluster $k$. We then concatenate all clustered memories $\{M^c_k\}_{k=1}^K$ and append them to the instruction $x$. We have in total $K \times D_m$ memory tokens after applying our clustering approach. We illustrate our approach in Figure~\ref{fig:diagram}.
We take the combination of processed memories and instruction $x$ as input to the LLM and train the embedding and LoRA parameters using cross-entropy loss, comparing with the ground-truth response $y$.

\section{Experiments}

\begin{table*}[h]
\setlength{\tabcolsep}{9pt}
\begin{center}
{
\resizebox{\textwidth}{!}{
\begin{tabular}{lccccc}
\toprule
 & \textbf{Number of memory} & \multicolumn{4}{c}{\textbf{ROUGE-L score} ($\uparrow$, \%)} \\
 \cline{3-6}
\textbf{Method} & \textbf{tokens ($\downarrow$)} & \textbf{Qwen2.5 1.5B} & \textbf{Gemma3 1B} & \textbf{StableLM2 1.6B} & \textbf{Average} \\
\midrule
Mean & \phantom{0,}128 & 12.79 & 12.37 & 12.05 & 12.40 \\
Concat & 1,024 & 15.34 & 13.91 & 12.14 & 13.80 \\
Clustering (ours) & \phantom{0,}512 & 15.16 & 13.45 & 13.36 & \textbf{13.99} \\
\hdashline
Mean & \phantom{0,}128 & 45.18 & 43.67 & 40.53 & 43.13 \\
Concat & 1,024 & 44.65 & 44.69 & 42.04 & 43.79 \\
Clustering (ours) & \phantom{0,}512 & 45.22 & 44.15 & 43.58 & \textbf{44.32} \\
\hdashline
Mean & \phantom{0,}128 & 43.59 & 22.35 & 46.39 & 37.44 \\
Concat & 1,024 & 37.24 & 25.75 & 46.29 & 36.42 \\
Clustering (ours) & \phantom{0,}512 & 44.27 & 24.35 & 51.59 & \textbf{40.07} \\
\bottomrule
\end{tabular}
}}
\end{center}
\caption{\textbf{
Main results across various models and datasets.} Personalized news headline generation at the top, followed by personalized tweet paraphrasing and personalized movie tagging.}
\label{tab:main}
\end{table*}

\begin{table*}[h!]
\setlength{\tabcolsep}{9pt}
\begin{center}
{\resizebox{\textwidth}{!}{
\begin{tabular}{lccccc}
\toprule
 & \textbf{Number of memory} & \multicolumn{4}{c}{\textbf{ROUGE-L score} ($\uparrow$, \%)} \\
 \cline{3-6}
\textbf{Method} & \textbf{tokens ($\downarrow$)} & \textbf{Qwen2.5 1.5B} & \textbf{Gemma3 1B} & \textbf{StableLM2 1.6B} & \textbf{Average} \\
\midrule
Mean & 128 & 12.79 & 12.37 & 12.05 & 12.40 \\
Clustering (ours) & 128 & 14.76 & 13.54 & 12.11 & \textbf{13.47} \\
\hdashline
Mean & 128 & 45.18 & 43.67 & 40.53 & 43.13 \\
Clustering (ours) & 128 & 44.34 & 43.69 & 42.90 & \textbf{43.64} \\
\hdashline
Mean & 128 & 43.59 & 22.35 & 46.39 & 37.44 \\
Clustering (ours) & 128 & 45.23 & 25.70 & 45.91 & \textbf{38.95} \\
\bottomrule
\end{tabular}
}}
\end{center}
\caption{\textbf{
Matching total number of memory tokens---results across various models and datasets.} Personalized news headline generation at the top, followed by personalized tweet paraphrasing and personalized movie tagging.}
\label{tab:matched}
\end{table*}

\subsection{Setup}
We use personalized news headline generation, tweet paraphrasing and movie tagging tasks from the LaMP benchmark \cite{salemi2023lamp}. All results are reported on the test set. We use Qwen2.5 1.5B \cite{qwen2}, Gemma3 1B \cite{team2025gemma}, StableLM2 1.6B \cite{bellagente2024stable} models, in all cases instruction-tuned. We use 8 texts in the memory set, 128 memory tokens per memory (with a compression rate of 4), and 4 clusters in the default case. Compression is performed via the selected frozen model with LoRA parameters fine-tuned for converting input texts into the given number of memory tokens. The LoRAs have a rank of 128, parameter $\alpha$ of 32, and dropout of 0.05. The fine-tuning is performed for 20 epochs with Adam optimizer and learning rate of $5\times 10^{-5}$. When there are more than 8 personalization texts available for the given examples, we use the closest 8 texts as selected via the BM25 approach. We evaluate the different approaches using the popular ROUGE-L score \cite{lin2004rouge}, which is a measure of overlap between the generated text and the ground-truth text. The LaMP benchmark uses ROUGE-L as an evaluation criterion \cite{salemi2023lamp} for our tasks, and its choice is motivated by personal style being reflected in the selection of words in the target text.

\subsection{Main Results}
We compare our clustering approach with the mean and concatenation baselines in Table~\ref{tab:main}. The results indicate that overall our solution leads to the best performance, on all considered tasks, while requiring significantly fewer memory tokens than the concatenation solution. We highlight that a key contribution of our method is it significantly improves efficiency of earlier best-performing solution (concat), while also improving over its performance. We only use 50\% of memory tokens used by the concatenation approach. We note our approach allows us to control the number of memory tokens by adjusting the number of clusters.

\subsection{Analyses}

\keypoint{Matching total number of memory tokens} To further analyse the benefits of our solution, we compare the performance of our solution with the mean baseline under the same total number of memory tokens. The mean approach uses 128 memory tokens, which is matched by using 4 clusters and 32 tokens for each memory in our clustering approach. The results in Table~\ref{tab:matched} confirm the usefulness of our solution.

The following three analyses utilize the Qwen2.5 1.5B model (unless otherwise mentioned) and personalized news headline generation tasks.

\begin{figure}[h!]
\vskip 0.2in
\begin{center}
\centerline{\includegraphics[width=\columnwidth]{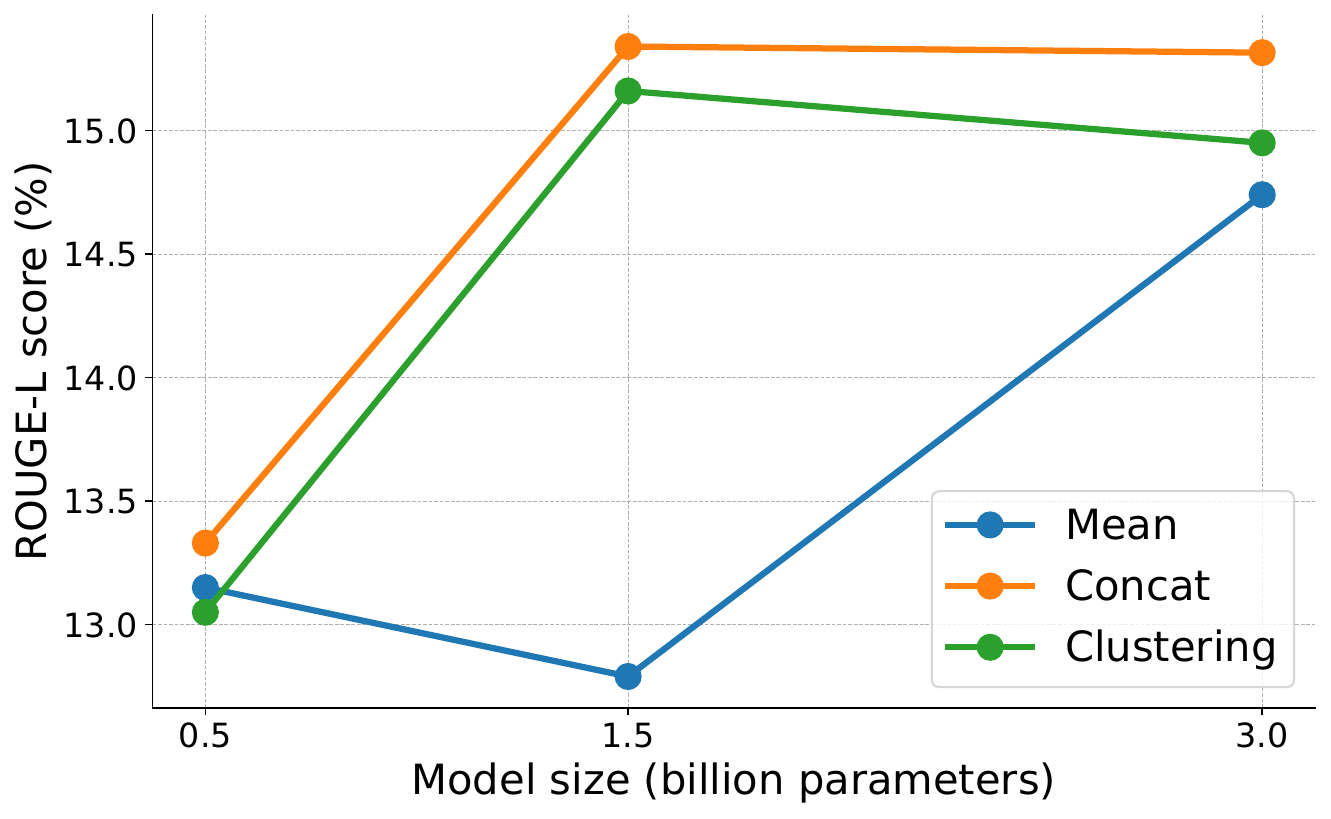}}
\caption{\textbf{Effect of model size on performance.} Larger models achieve better performance, with consistent behaviour across the different approaches.
}
\label{fig:model_size_rouge}
\end{center}
\vskip -0.2in
\end{figure}

\keypoint{Model size} We utilize the Qwen2.5 model in three on-device sizes: 0.5B, 1.5B, and 3B. Our analysis in Figure~\ref{fig:model_size_rouge} shows that larger models achieve better performance, with concatenation and our clustering approaches offering the best performance.

\begin{figure}[h!]
\vskip 0.2in
\begin{center}
\centerline{\includegraphics[width=\columnwidth]{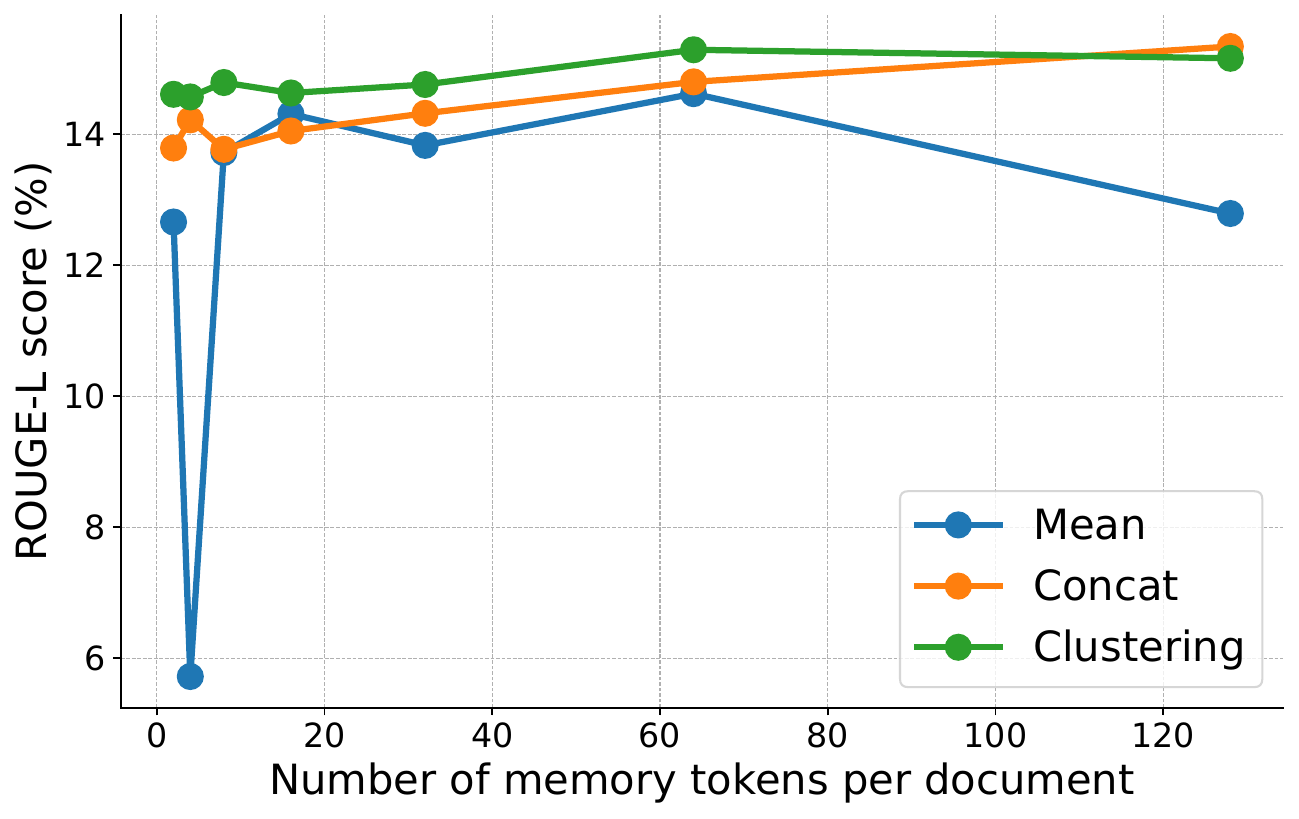}}
\caption{\textbf{Effect of the number of memory tokens per document on performance.} More memory tokens typically lead to better performance, but good performance can also be achieved with fewer memory tokens. Our clustering solution achieves the best performance in most cases.
}
\label{fig:memory_tokens_rouge}
\end{center}
\vskip -0.2in
\end{figure}

\keypoint{Number of memory tokens} We study the impact of the number of memory tokens per document on performance in Figure~\ref{fig:memory_tokens_rouge}. We see that more memory tokens typically lead to better performance. However, good performance can also be achieved with fewer memory tokens. The results also show that our clustering solution achieves the best performance in most cases.

\begin{figure}[h!]
\vskip 0.2in
\begin{center}
\centerline{\includegraphics[width=\columnwidth]{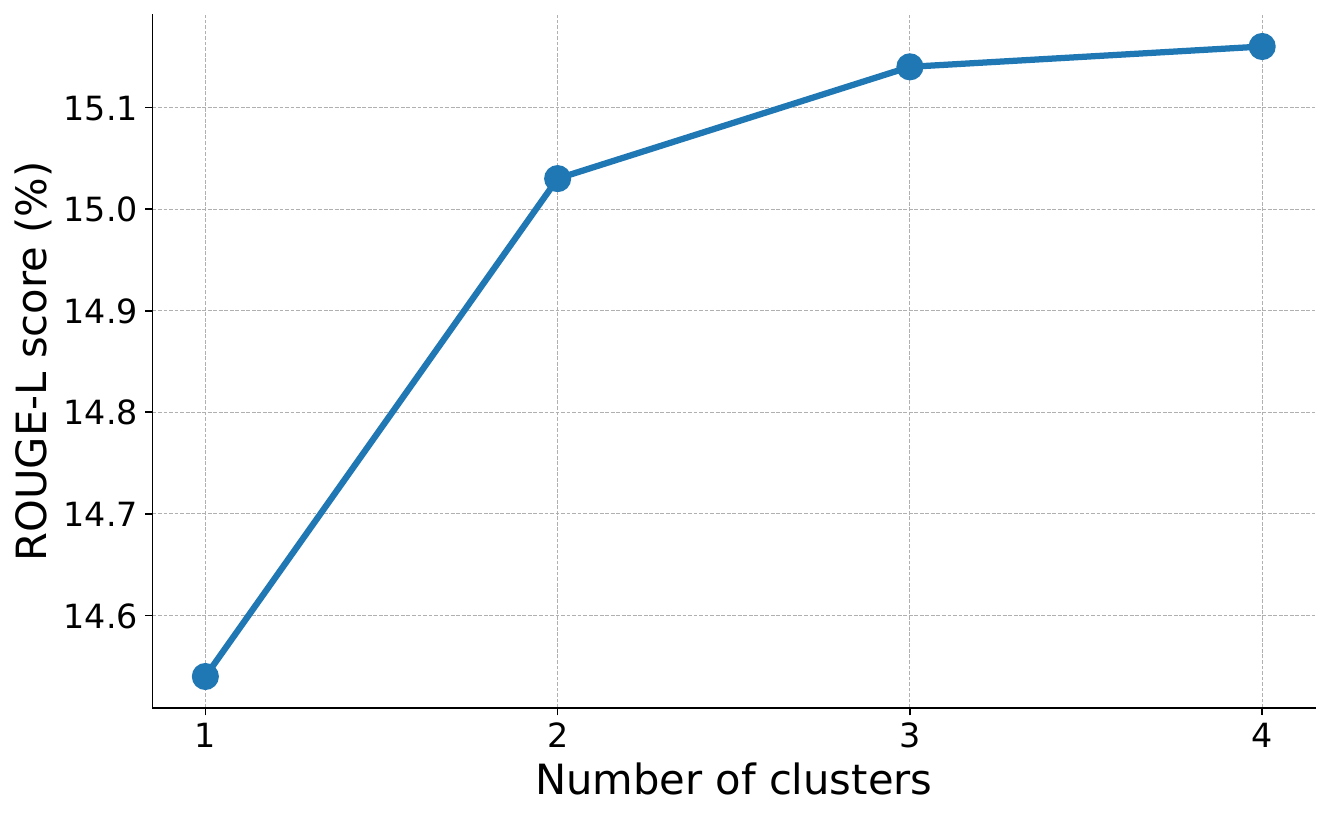}}
\caption{\textbf{Effect of the number of clusters on performance.} Using more clusters in our solution improves performance.
}
\label{fig:clusters_rouge}
\end{center}
\vskip -0.2in
\end{figure}

\keypoint{Number of clusters} Figure~\ref{fig:clusters_rouge} shows that using a higher number of clusters is beneficial for improving the performance, even though the required number of memory tokens scales linearly by the number of clusters. The performance drops are relatively small when using three or two clusters instead of four, but the decrease is significant when using only one cluster.

\section{Conclusion}
We have introduced an efficient method to personalize on-device LLMs via clustering-driven memory compression. The method leads to improved performance compared to both concatenation and computing the mean of memories, while using fewer memory tokens than concatenation. Analyses across diverse settings have confirmed the usefulness of the solution.

\bibliographystyle{IEEEbib}
{\small\bibliography{refs}}

\end{document}